\documentclass[letterpaper, 10 pt, conference]{ieeeconf}  % Comment this line out if you need a4paper

\IEEEoverridecommandlockouts                              % This command is only needed if 
\usepackage{graphicx}
\graphicspath{{./figures/glove_psm_tf/}}

 \usepackage{grffile} 
\usepackage{epsfig} % for postscript graphics files
\usepackage{mathptmx} % assumes new font selection scheme installed
\usepackage{times} % assumes new font selection scheme installed
\usepackage{amsmath} % assumes amsmath package installed
\usepackage{amssymb}  % assumes amsmath package installed

\usepackage{subcaption}
\usepackage{multirow}
\usepackage{hyperref}
\usepackage{xcolor}
\usepackage{cite}
\usepackage{colortbl}
\usepackage{dirtytalk}
\usepackage[normalem]{ulem}

\usepackage[ruled,linesnumbered]{algorithm2e}
\usepackage{svg}

\usepackage{array}
\newcolumntype{P}[1]{>{\centering\arraybackslash}p{#1}}
\newcolumntype{M}[1]{>{\centering\arraybackslash}m{#1}}

\newcommand{\gn}[1]{\textbf{#1}}

\title{\LARGE \bf
Sensory Glove-Based Surgical Robot User Interface
}

\author{Leonardo Borgioli$^{1*}$, Ki-Hwan Oh$^{1*}$, Valentina Valle$^{2}$, Alvaro Ducas$^{2}$, Mohammad Halloum$^{2}$,\\ Diego Federico Mendoza Medina$^{2}$, Arman Sharifi$^{2}$, Paula A L\'opez$^{2}$, Jessica Cassiani$^{2}$,\\ Milo\v s \v Zefran$^{1}$, Liaohai Chen$^{2}$ and Pier Cristoforo Giulianotti$^{2}$% <-this % stops a space
\thanks{$^{*}$First two authors contributed equally to this work.}
\thanks{$^{1}$Robotics Lab,  Department of Electrical and Computer Engineering, University of Illinois Chicago, Chicago, IL 60607, USA.}%
\thanks{$^{2}$Surgical Innovation and Training Lab,  Department of Surgery, University of Illinois Chicago, Chicago, IL 60607, USA.}%
}

\begin{document}

\maketitle
\thispagestyle{empty}
\pagestyle{empty}

%%%%%%%%%%%%%%%%%%%%%%%%%%%%%%%%%%%%%%%%%%%%%%%%%%%%%%%%%%%%%%%%%%%%%%%%%%%%%%%%
\begin{abstract}

Robotic surgery has reached a high level of maturity and has become an integral part of standard surgical care. However, existing surgeon consoles are bulky, take up valuable space in the operating room, make surgical team coordination challenging, and their proprietary nature makes it difficult to take advantage of recent technological advances, especially in virtual and augmented reality. One potential area for further improvement is the integration of modern sensory gloves into robotic platforms, allowing surgeons to control robotic arms intuitively with their hand movements. We propose one such system that combines an HTC Vive tracker, a Manus Meta Prime 3 XR sensory glove, and SCOPEYE wireless smart glasses. The system controls one arm of a da Vinci surgical robot. In addition to moving the arm, the surgeon can use fingers to control the end-effector of the surgical instrument. Hand gestures are used to implement clutching and similar functions. In particular, we introduce clutching of the instrument orientation, a functionality unavailable in the da Vinci system. The vibrotactile elements of the glove are used to provide feedback to the user when gesture commands are invoked. A qualitative and quantitative evaluation has been conducted that compares the current device with the dVRK console. The system is shown to have excellent tracking accuracy, and the new interface allows surgeons to perform common surgical training tasks with minimal practice efficiently. 

% this suggests that the interface is highly intuitive. The proposed system is inexpensive, allows rapid prototyping, and opens opportunities for further innovations in the design of surgical robot interfaces.

%v3 The evolution of robotic surgical devices has led to advancements in precision, dexterity, and user interaction. Most of the current surgical robots are controlled by the surgeons sitting in a console with manipulators and stereo vision included. However, the current surgical robot console has some limitations in modern Operation Rooms (ORs) due to the large consumption of space and lack of additional feedback beyond visual cues. This paper proposes a novel surgeon-robot interface using a sensory glove imitating the manipulators in the console with additional features through tracking the gesture of the hand. An improved clutch mechanism enhances a more flexible control of the robot arms, while vibrotactile feedback offers additional feedback to the surgeons. Implementing this architecture addresses space constraints and the need for enhanced feedback, aiming to improve surgical outcomes and user experience in robotic-assisted procedures.

\end{abstract}

% The bulky design exacerbates spatial constraints\cite{kanji2021room}, creating possible issues during the procedure \cite{cofran2021barriers}.

%%%%%%%%%%%%%%%%%%%%%%%%%%%%%%%%%%%%%%%%%%%%%%%%%%%%%%%%%%%%%%%%%%%%%%%%%%%%%%%%

\section{INTRODUCTION}

Robotic surgery has reached a high level of maturity and has become an integral part of standard surgical care. For the surgeon, it offers increased precision and dexterity through efficient interaction interfaces. However, despite its high acceptance and success, it faces challenges in modern operating rooms (ORs). The surgeon consoles are bulky and take up valuable space~\cite{kanji2021room}. A redesigned workspace promises to improve turnaround time, increase throughput, and reduce cost~\cite{bharathan2013operating}. Studies further emphasize the potential of ergonomic workspace design and nonoperative process redesign to advance the efficiency of robotic surgical devices~\cite{sandberg2005deliberate, stahl2006reorganizing}. Another challenge with current consoles is that the surgeon is separated from the rest of the team, which can lead to challenges in communication~\cite{simorov_review_2012} and can impact the efficiency of the operational workflow in tasks such as robotic docking~\cite{cofran2021barriers}. Finally, the proprietary nature of existing consoles makes it challenging to innovate in this space despite rapid advances in fields such as augmented and virtual reality that could benefit the surgeon.

Several studies have attempted to address some of these shortcomings. In \cite{hong_head-mounted_2019}, a novel method is proposed to control the endoscope through a head-mounted interface (HMI) integrated with a surgical robot system. And \cite{wen_hand_2014, Ai2024} control the surgical robot arm and the Augmented Reality (AR) interface through various hand gestures.

One potential area for further improvement and technological innovation in robotic surgery is the integration of modern sensory gloves into robotic platforms, which allow surgeons to intuitively control robotic arms with their hand movements without the need for additional manipulators that are customary for existing consoles. Furthermore, most end-effectors of surgical robot instruments, such as the Fenestrated Bipolar Forceps (FBF) and the Large Needle Driver (LND), are limited to the use of opposing blades ~\cite{intuitive2023catalog} due to the design of the console manipulators (as the last joint is controlled only by pinching two fingers); glove-based control could offer the surgeon additional dexterity while using these instruments.
%\textcolor{red}{also enhaced surgical precision, patient safety, and medical training by adding new form of sensorization \cite{salvadores2023sensorised}}.

Glove-based control is intuitive and is popular in robotics applications outside the surgical field. For instance, a low-cost wireless system with a glove equipped with bending sensors is described in~\cite{mercilinraajini2023mem}. In~\cite{burns2020design}, the authors demonstrate glove-based control in rehabilitation, using gloves equipped with Inertial Measurement Units (IMUs) for patient exercises. Glove-based control for telemanipulation tasks, integrating force sensors for enhanced feedback, is explored in~\cite{brygo2017synergy}. These studies explore various sensor modalities and develop communication techniques and control algorithms to achieve sophisticated and effective glove-based control systems for robotic arms. 

In our work, we develop a system in which an HTC Vive Tracker~\cite{htcvive} is attached to a Manus Meta Prime 3 Haptic XR sensory glove~\cite{manus}, equipped with multiple IMUs and bending sensors for finger tracking to control one arm of the da Vinci surgical robot paired with the da Vinci Research Kit (dVRK)~\cite{dvrk}. Visual feedback to the surgeon is provided through SCOPEYE wireless smart glasses~\cite{scopeye}. In addition, this setup manipulates the robotic arm's end effector (jaw of the instrument) through finger movements. Moreover, we implemented hand gesture recognition to substitute pedal inputs of the current da Vinci console (e.g., clutch). One of the gestures is employed to clutch the instrument orientation, a functionality not provided by the da Vinci system. The glove has vibrotactile sensors to inform the user that a gesture has been activated.

%\clearpage %to keep to avoid spacing errors

We evaluated the proposed system by having five surgeons and two medical doctors (no completed residency in surgery) perform a modified version of a robot-assisted surgical exercise using our system and the dVRK console. The system has been evaluated in terms of tracking accuracy and movement efficiency. After completing the exercise, the participants completed the System Usability Scale~\cite{SUS} and the NASA Task Load Index~\cite{nasa_questionnaire} questionnaires to measure the usability and workload of our proposed architecture.
These evaluations show that the system is responsive and that surgeons can effectively complete training exercises with minimal practice with the new interface, suggesting that the interface is highly intuitive. The proposed system is inexpensive, as it uses off-the-shelf components, and it dramatically reduces space requirements. Our work also demonstrates how modalities such as vibrotactile feedback can efficiently provide additional information to the surgeon without cluttering the visual field. Finally, the proposed architecture allows for rapid prototyping and opens opportunities for further innovations in the design of surgical robot interfaces. 

\section{METHODOLOGY}

\begin{figure}[t]
    \centering
    \includegraphics[ width=\linewidth]{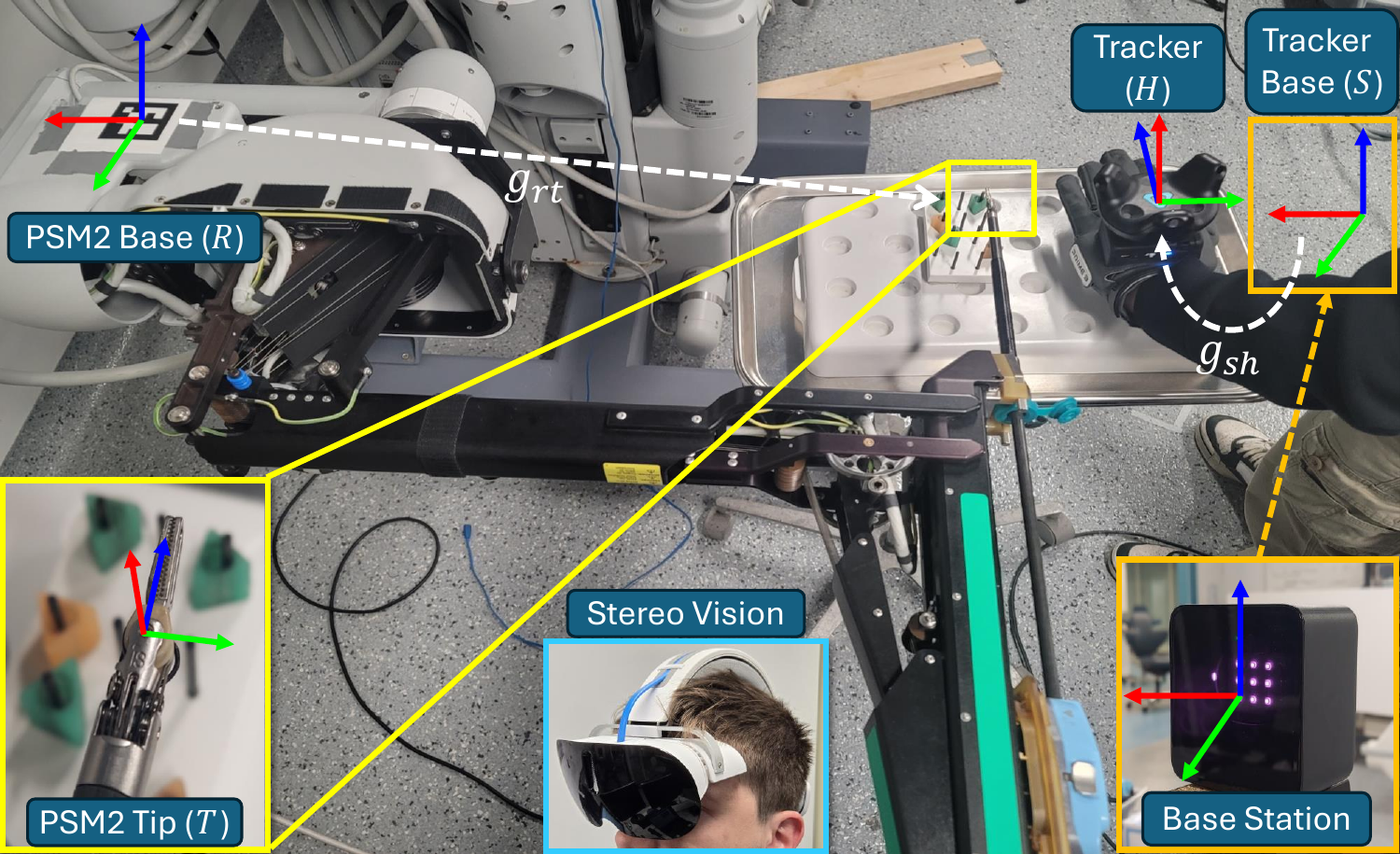}
    \caption{The setup showing the configuration of the da Vinci arm (PSM) and the Manus glove with the HTC Vive Tracker attached. The user wears the SCOPEYE smart glasses for the stereo endoscopic view. Shown are the $x$ (red), $y$ (green), and $z$ (blue) axes of the relevant coordinate frames. }
    %The tracker's base frame is established using its initial position captured from the Base Station and has the same orientation as the PSM base frame.}
    \label{fig:arm_glove_setup}
    \vspace{-5mm}
\end{figure}

\subsection{System Architecture}
% The objective of this work is therefore to be able to mimic the dVRK's console controllers with a simpler device like the Manus Meta VR glove.
The system integrates several components. The surgical robot platform comprises the first-generation da Vinci robot paired with the dVRK. 
One of the Patient Side Manipulators (PSMs) is equipped with the Fenestrated Bipolar Forceps (FBF) instrument and is subject to glove control. The arm is configured as in Fig.~\ref{fig:arm_glove_setup}. The HTC Vive Tracker attached to the Manus Meta Prime 3 XR glove provides the pose of the hand, while the fingers are tracked by the glove. Instead of using image-based hand-tracking devices as in~\cite{Ai2024}. We decided to use sensory gloves because video-based hand-tracking (commonly used in AR, VR, and Leap Motion applications) performs poorly in the presence of visual occlusion, background noise, or variations in room lighting~\cite{korayem2021controlling}. These occur frequently in the OR, such as when other surgical team members use secondary screens to look at details in the scene. Moreover, sensory gloves offer vibrotactile feedback. In addition, classifier models were trained for hand gesture recognition; hand gestures were implemented to substitute the da Vinci console's pedal inputs. SCOPEYE smart glasses that provide the same visual feedback as the High-Resolution Stereo Video (HRSV) of the da Vinci console and have a maximum latency of 7ms~\cite{scopeye} were used to further simplify the setup.

The system runs the Robot Operating System (ROS)~\cite{ros} that facilitates seamless communication and coordination between the Manus glove, the HTC Vive Tracker, and dVRK, ensuring real-time data streaming and responsive interaction between the user and the control interface. 
% and can be used by developers to develop We aimed to construct a more intuitive command pipeline, offering a new form of feedback that can be utilized by developers

\subsection{da Vinci Arm Kinematics}

For the full da Vinci system, the kinematic chain from the Patient Side Manipulator (PSM) tip to the Endoscope Camera Manipulator (ECM) tip involves Setup Joints (SUJs). The poses provided by the dVRK exhibit errors within a range of $\pm 5cm$ for translation and fall between $5\sim10$ degrees for orientation. These inaccuracies arise due to challenges in calibrating the SUJs precisely, as reported in~\cite{dvrk}. Consequently, we implemented a custom calibration method for the dVRK using fiducial markers~\cite{oh2023framework} that eliminates the influence of SUJs and allows us to accurately determine the transformation $g_{rt}$ between the PSM base (frame $R$) and the instrument tip (frame $T$), see Fig.~\ref{fig:arm_glove_setup}.

\subsection{Scaling Workspace}
\label{sec:scale_ws}

To ensure safety for both the glove and the instrument tip, we define their operational workspaces as cubes, with a larger side $L_h$ (30cm) for the glove and a side $L_t$ (7.5cm) for the instrument tip, centered at their initial positions. In both cases, movements beyond the cube volume are projected to the cube surface.

The glove movements tend to be significantly larger than the desired movements of the PSM tip ($L_h > L_t$). The motion of the glove is thus scaled to match the range needed for PSM operations:
\begin{equation}
\label{eq:rescale}
\Delta p_t = \eta \, \Delta p_h
\end{equation}
where $\Delta p_h$ represents the translational movement of the glove from its initial position, $\Delta p_t$ denotes the expected translational movement of the PSM from its initial position, and $\eta={L_t}/{L_h}$ is the scaling factor. Similarly to the existing surgeon console, the scaling factor $\eta$ can be adjusted.
%This rescaling ensures the glove's broader movements are precisely converted into the PSM's required action.

\subsection{Glove Kinematics}

To monitor the glove's position and orientation within the environment, we attached a HTC Vive Tracker to it. The glove position and orientation data are updated at 120 Hz using the data from the tracker's IMUs and HTC Vive Base Stations~\cite{basestation, Htc}. 

Tracking of the relative transformation between the fingers and the wrist is accomplished using the five sets of flexible sensors with 2 Degrees of Freedom (DoF) and six sets of 9 DoF IMUs of the Manus glove, operating at 90Hz with $\pm 2.5^{\circ}$ accuracy~\cite{manus}. The glove software provides a hand skeleton mesh inspired by the Mediapipe framework~\cite{bazarevsky2019mediapipe}. This mesh is dynamically adjusted based on the readings from the glove sensors, allowing for a real-time representation of the hand pose.

\begin{table}[!t]
% \vspace{-2mm}
    \centering
    \begin{tabular}{M{1.1cm}|c|M{1.1cm}|c}
    \hline \hline
        \textbf{Hand Gesture} & \textbf{Representation} &\textbf{Hand Gesture}& \textbf{Representation} \\
    \hline
%        \textbf{None} & \begin{minipage}{0.1\textwidth} \includegraphics[angle=-90, origin=c, width=\columnwidth]{figures/gestures/none.jpg} \end{minipage} &
        \hline
        \textbf{Ring} & \begin{minipage}{0.1\textwidth} \includegraphics[angle=-90, origin=c, width=\columnwidth]{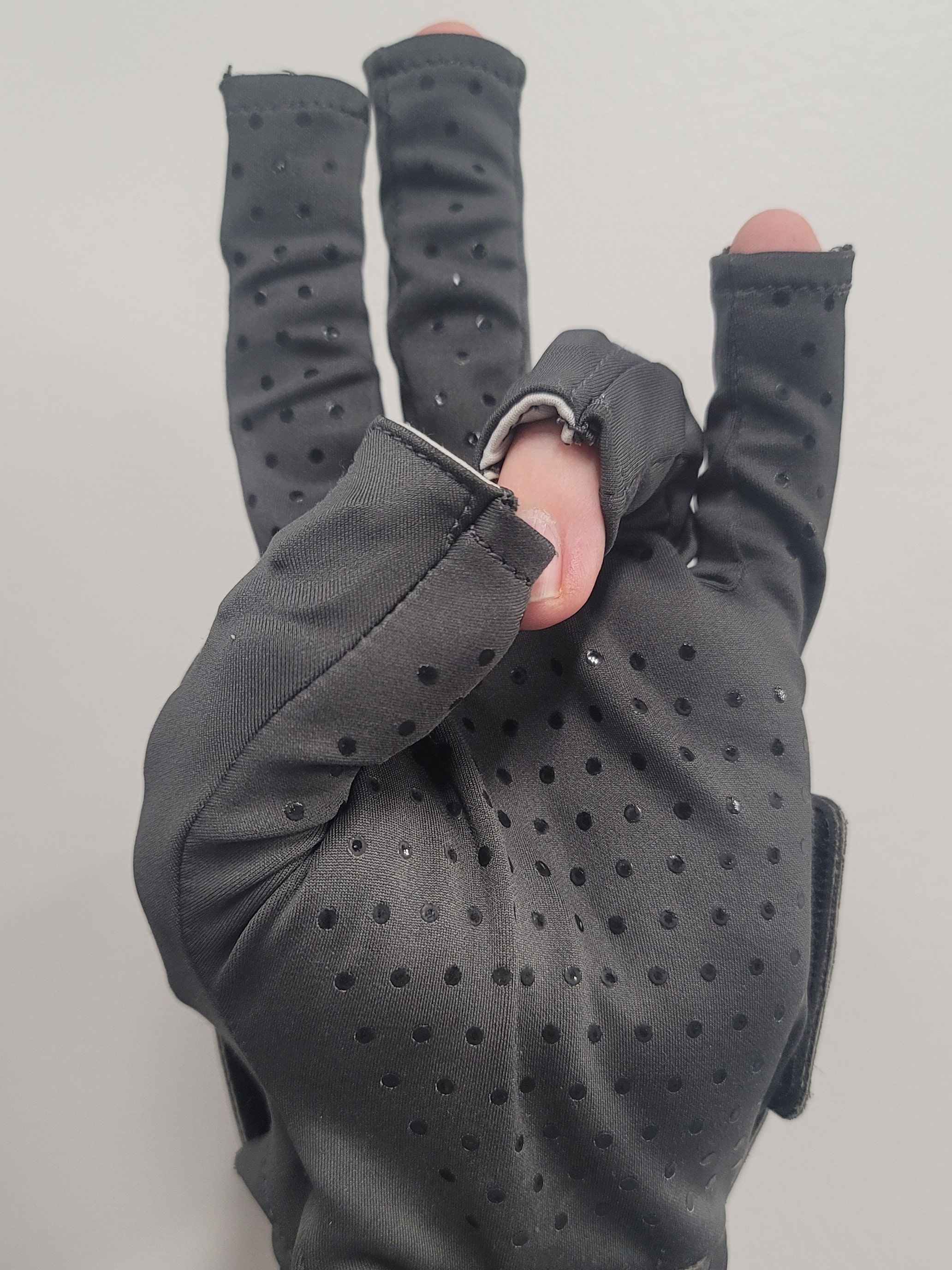} \end{minipage} &
        \textbf{Clutch} & \begin{minipage}{0.1\textwidth} \includegraphics[angle=0, origin=c, width=\columnwidth]{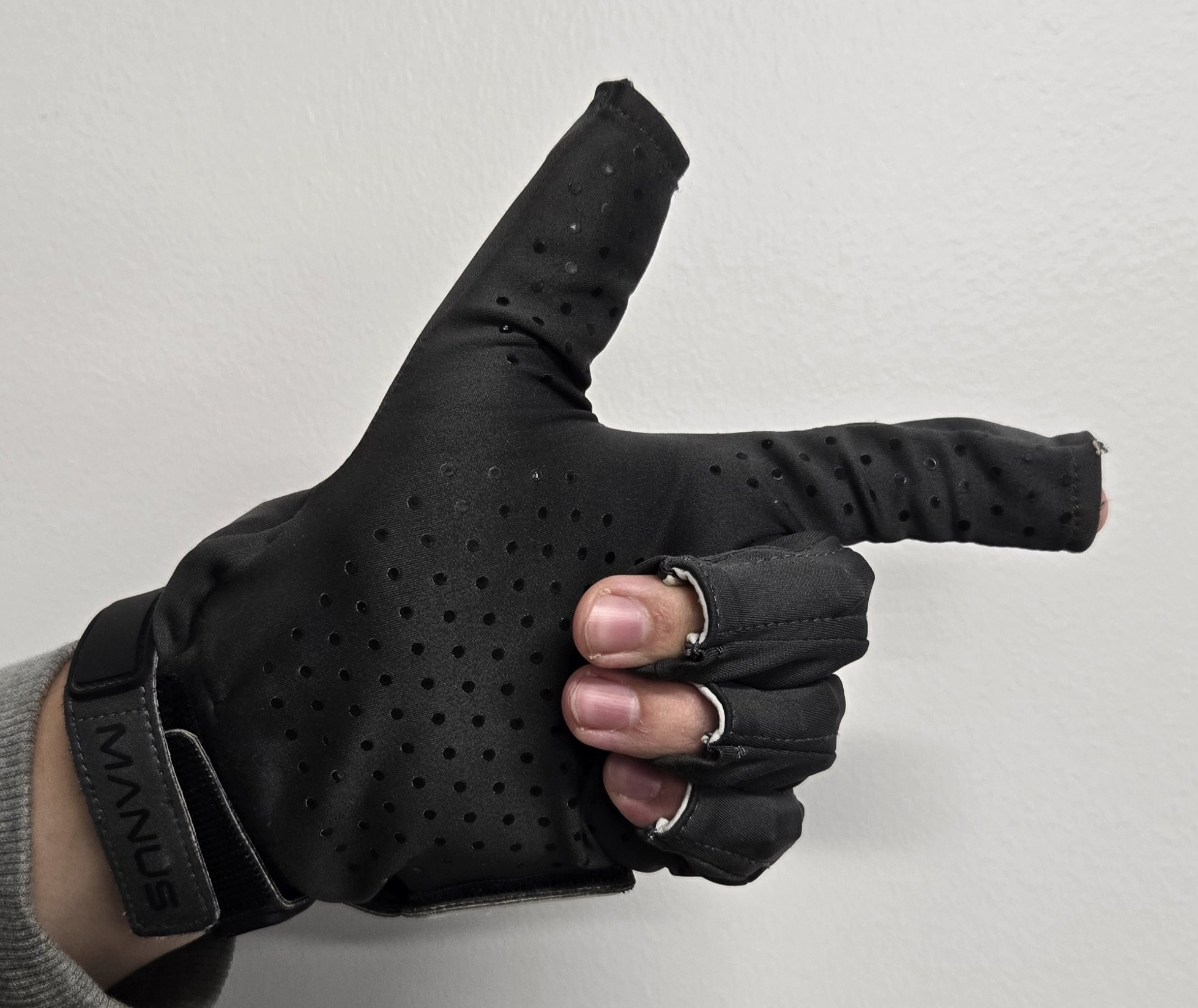} \end{minipage} \\
        \hline
%        \textbf{Thumbs up} & \begin{minipage}{0.1\textwidth} \includegraphics[angle=90, origin=c,width=\columnwidth]{figures/gestures/thumb_up.jpg} \end{minipage} & \multicolumn{2}{c}{}\\
        \hline
    \end{tabular}
\caption{Gestures trained for distinct functionalities. \textbf{Clutch} invokes the clutch mechanism and \textbf{Ring} turns the robot control on or off. If neither of these two gestures is recognized, the system returns \textbf{None}.}
\label{tab:glove_gestures}
\vspace{-7mm}
\end{table}

The HTC Vive Tracker and Base Station coordinate frames are depicted in Fig.~\ref{fig:arm_glove_setup}. At initialization, the tracker frame $H$ and the tracker base frame $S$ coincide and are aligned with the PSM instrument tip frame $T$. The desired translation for the PSM end-effector is calculated by scaling the displacement of $H$ from $S$ as described above to create an intuitive control for the user.

For the goal orientation of the PSM end effector, we determine the relative orientation between the metacarpophalangeal (MP) and distal interphalangeal (DIP) joint of the sensory glove's index finger and add this rotation to the one provided by the HTC Vive Tracker readings. This ensures that the system follows the index finger of the user, making the system more intuitive and reducing user fatigue.

In our architecture, the Euclidean distance between the tips of the index and thumb fingers was used to control the jaw of the end effector. To map the distance of the fingers to the distance between the tips of the jaw, we use an expression analogous to Eq.~\eqref{eq:rescale}. The maximum opening of the jaw is achieved when the distance between the two fingertips is $10$cm. This value has been selected as it is the $95\%$ percentile of the length of the index finger of women~\cite{hand_percentile}, ensuring that all users can precisely and intuitively control the instrument jaw within the full range of motion.

% \subsection{Finger Tracking}

% It is important to note that we chose this approach instead of building a full kinematic chain between the sensory glove and the HTC Vive tracker due to the lack of rigidity of the glove/tracker connection.

\subsection{Hand Gesture Recognition}

Hand gestures serve as a seamless and intuitive means of interaction and open up the possibility of introducing a wide variety of inputs surpassing what the current consoles can provide. For example, \cite{wen_hand_2014} shows that it is possible to implement 8 different commands.

In our work, we train 4 different multiclassification models for hand gesture recognition using our custom dataset. Our implemented architecture can currently distinguish between three distinct gestures: \textbf{None}, \textbf{Ring} (ring finger and thumb in contact), and \textbf{Clutch} (middle, ring, and pinky fingers closed) as illustrated in Table~\ref{tab:glove_gestures}. \textbf{Clutch} activates the clutch mechanism described in Section~\ref{sec:clutch}. Holding the \textbf{Ring} gesture for 2 seconds initiates/suspends robot tracking. This implementation is open to allow new gestures to be added with a relatively small dataset. Further details are provided in Section~\ref{Sect:Gestures}.

\subsection{Clutch}
\label{sec:clutch}
\begin{table}[!t]
    \centering
    \begin{tabular}{c|M{1.1cm}|c|M{1.1cm}}
    \hline \hline
        \textbf{Input Hand Pose} & \textbf{Predicted Gesture} & \textbf{Input Hand Pose} & \textbf{Predicted Gesture}\\
    \hline
        % \begin{minipage}{0.10\textwidth} \includegraphics[trim={7cm 5cm 5cm 7cm}, clip, width=\columnwidth]{figures/gestures/none_prediction.png} \end{minipage} & None
        \begin{minipage}{0.10\textwidth} \includegraphics[trim={5cm 5cm 3cm 3cm}, clip, width=\columnwidth]{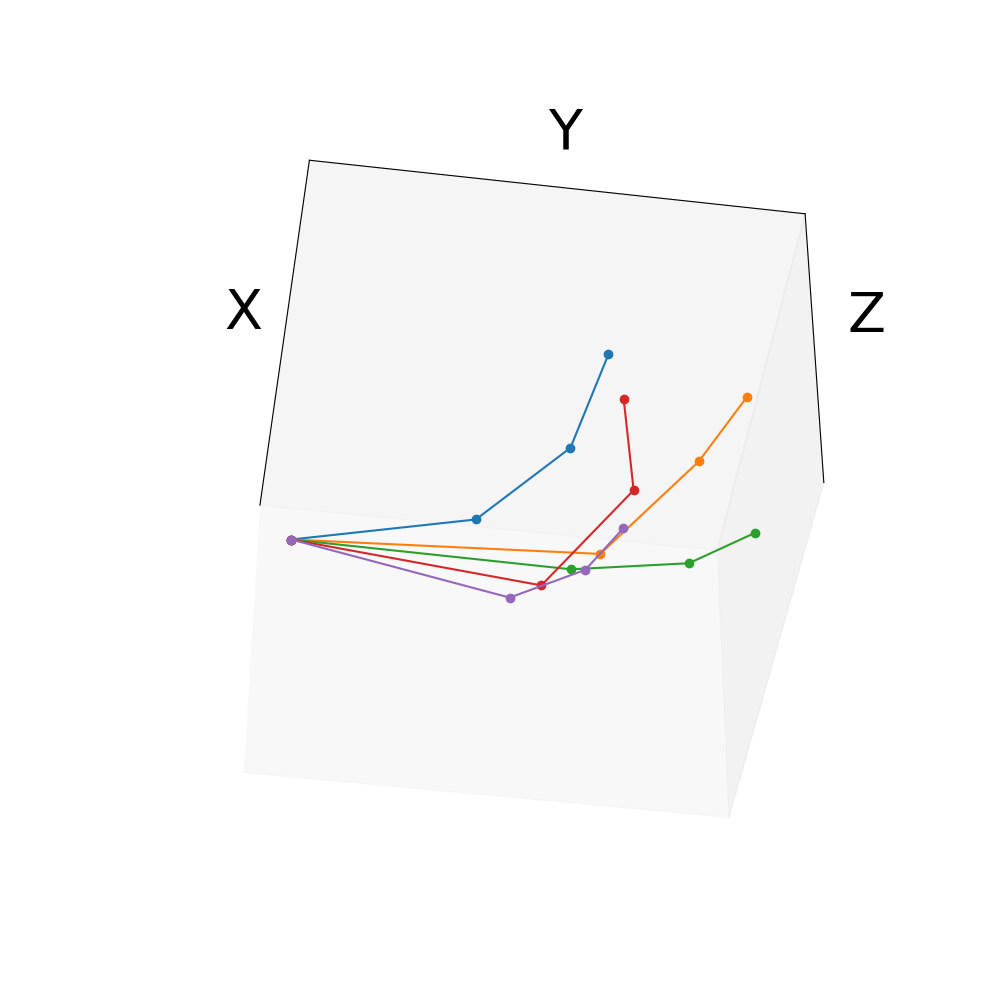} \end{minipage} & \textbf{Ring} 
        &\begin{minipage}{0.10\textwidth} \includegraphics[trim={2cm 2cm 2cm 2cm}, clip, width=\columnwidth]{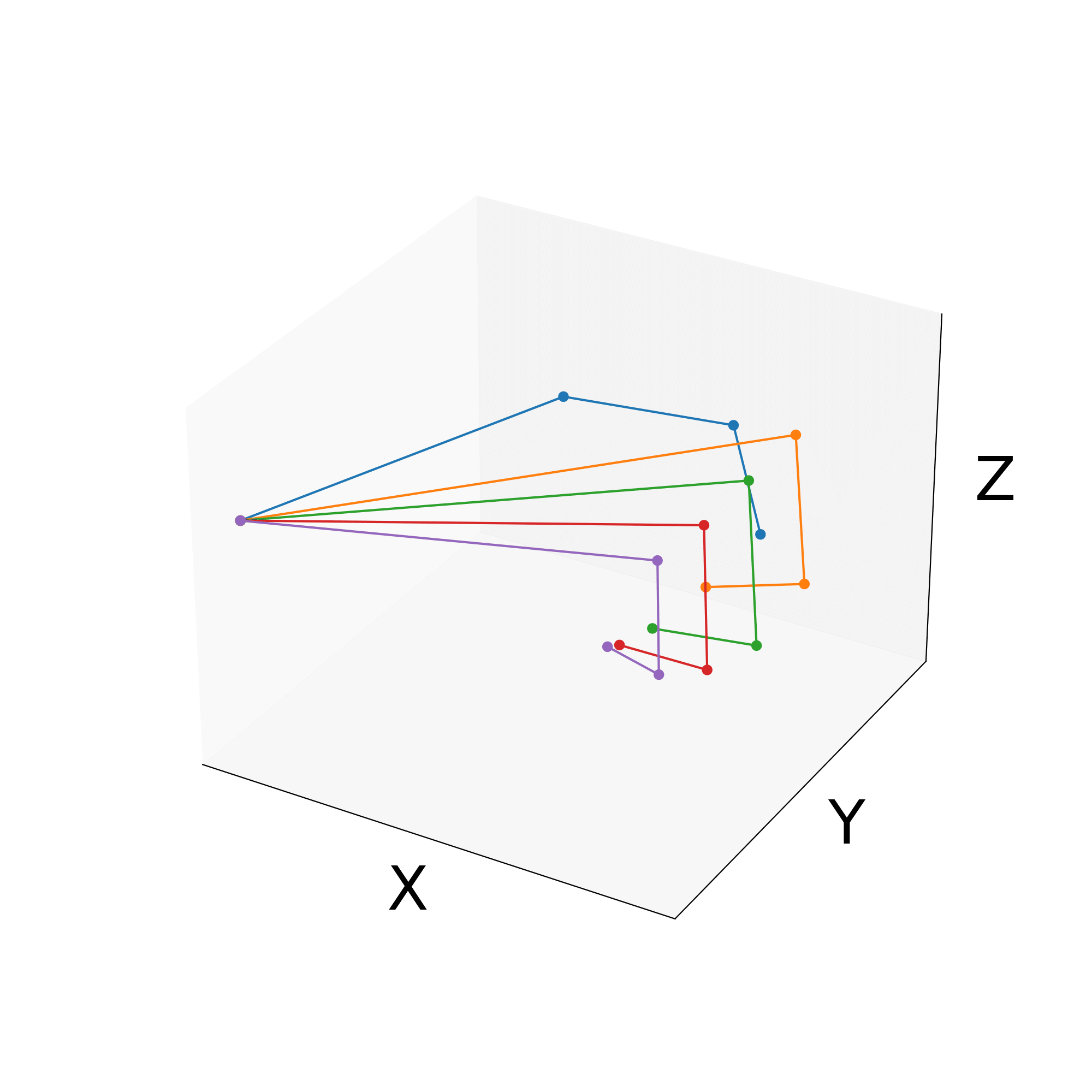} \end{minipage} & \textbf{Clutch}\\ \hline\hline
    \end{tabular}
\caption{Examples of point clouds used as input}% for gesture prediction.}
\label{tab:glove_pred}
\vspace{-5mm}
\end{table}

%DO NOT DELETE THE SPACES (they align the images on the next page)

The clutch functionality in the da Vinci surgical robot serves as a safety feature and a means of control for the surgeon during robotic-assisted surgeries. This mechanism essentially allows the surgeon to recenter the workspace of the surgeon console by temporarily disengaging the movement of the robotic arms while still maintaining control of the Master Tool Manipulators (MTMs).

% When the surgeon operates the da Vinci system's master console, which consists of hand and finger controls, they can engage or disengage the clutch as needed. This allows them to stop the movement of the robotic arms,  enabling them to make adjustments or reposition the controllers as required during the procedure.

In the dVRK system, the clutch is engaged with a foot pedal. However, the current rotation of the tool is maintained and cannot be changed by fixing the orientation of the controllers. Our system implements full clutching (both translation and rotation). The clutch is engaged by performing the \textbf{Clutch} gesture as described next.

The clutching mechanism includes 3 phases:

\begin{itemize}
    \item \textbf{Leading Edge:} When the \textbf{Clutch} gesture is detected, the clutch is engaged, activating the vibrotactile feedback of the glove. The current configuration of the PSM $g_{rt}^0$ is recorded, and the robot is prevented from moving.
    \item \textbf{Active Clutch:} The vibrotactile feedback continues, and the PSM is held steady. The user can change the pose of the glove to a comfortable configuration.
    \item \textbf{Trailing Edge:} When the gesture changes, the clutch is disengaged, and the vibrotactile feedback is deactivated. The tracker frame $H$ and the tracker base frame $S$ are aligned to the axes of the PSM tip frame $T$, whose configuration is described by $g_{rt}^0$.
\end{itemize}

% \noindent \textbf{Leading Edge:} When the \textbf{Clutch} gesture is detected, the clutch is engaged, activating the vibrotactile feedback of the glove. The current configuration of the PSM $g_{rt}^0$ is recorded, and the robot is prevented from moving. % Note that the jaw of the instrument is allowed to move.

% \noindent \textbf{Active Clutch:} The vibrotactile feedback continues to be activated. The PSM is held immobile. The glove can be moved to an arbitrary pose, allowing the user to move to a comfortable configuration.

% \noindent \textbf{Trailing Edge:} When the gesture becomes other than \textbf{Clutch}, the clutch is disengaged, and the vibrotactile feedback is deactivated. The tracker frame $H$ and the tracker base frame $S$ are aligned, and their axes are aligned with the axes of the PSM tip frame $T$, whose configuration is described by $g_{rt}^0$.

\section{Hand Gesture Recognition}
\label{Sect:Gestures}

We created a custom dataset for training hand gesture recognition, recording approximately $3-5$ minutes of the glove worn by different users performing one of the gestures in Table~\ref{tab:glove_gestures}, including the \gn{None}. The data consists of point clouds with 21 hand joint landmarks, capturing the relative orientations to the base, set when the glove is activated. The orientations are represented by quaternions with respect to the glove base, initialized when the glove is turned on, resulting in a total of 84 features. The translations were not added because they vary for the same hand posture with each launch. Using only rotations thus makes it easy to capture gestures of interest with only a few recordings. Table~\ref{tab:glove_pred} shows examples of point clouds for different gestures. Each recording contains one gesture, and the pose of the hand is changed to be as diverse as possible. 
We added $150,000$ gestures with sets of random normalized quaternions to the \gn{None} class to prevent the models from recognizing a random gesture as one of the gestures of interest (Ring or Fist). The final dataset consists of $15,557$ \gn{Fist} instances, $17,036$ \gn{Ring} instances, and $178,217$ \gn{None} instances. A simple one-hot encoding was applied due to the small number of classes. 

% We selected this approach because, although the base changes with each glove use, it effectively captures all possible hand rotations during recording. Each recording has been made while changing the pose of the hand as much as possible to ensure the widest variety of data. Each recording corresponds to one gesture (or no gesture), so it was easy to attribute a class to each sample when merging all the data. To augment the \gn{None} class, we generated  This ensures the correct proportionality, providing approximately ten times more data for the \gn{None} class; this quantity was determined by calculating the average ratio of class counts across two exercises.

% \input{tables/gesture_dataset}

\subsection{Classifiers}

We initially used Random Forests (RF)~\cite{breiman2001random} as a model for gesture recognition. Subsequently, we explored the Feedforward Neural Network (FNN)~\cite{Goodfellow-et-al-2016} classifiers analogous to those introduced in~\cite{rysbek2023recognizing}. However, the networks have been specifically tuned for classification. The structure of the FNN consists of an input layer ($n_{0}$) sized according to the input characteristics, followed by two fully connected layers ($n_{1}$, $n_{2}$), where the number of neurons is proportional to the size of the previous layer ($n_{1} = 2\cdot n_{0}$, $n_{2} = 2/3\cdot n_{1}$). The final layer ($n_{f}$) has $3$ output to function as a classifier, predicting one of the three possible classes.

In addition to these established models, we investigated the Light Gradient Boosting Machine (L-GBM)~\cite{NIPS2017_6449f44a}.% L-GBM employs a sequential approach in which decision trees are added successively, with each tree correcting errors from the ensemble of preceding trees. 
Unlike traditional gradient-boosting methods like XGBoost~\cite{XGBoost}, where trees grow depth-wise, L-GBM trees grow leaf-wise, making training faster. Tailored for efficient training on large datasets, they are particularly suitable for applications with resource constraints and can quickly provide good predictions.

We also trained a Histogram-Based Gradient Boosting (HGBT) model that combines traditional gradient boosting~\cite{friedman2002stochastic} and a histogram-based algorithm~\cite{NIPS2017_6449f44a}. The HGBT works with binned feature values rather than individual data points, which is the case with traditional gradient boosting methods. This approach accelerates the computation of information gained during tree construction by constructing histograms of input features. The binning process reduces the number of unique feature values, resulting in faster training times and decreased memory usage.

 To optimize the model hyperparameters, we adopted Grid-Search Cross-Validation (GridSearchCV)~\cite{GridCV}. First, we split the dataset into training and testing sets with the $8:2$ ratio. The search process used a 5-fold cross-validation strategy, in which each combination was evaluated across different data splits. The grid includes five hyperparameters with three different values, resulting in 1215 combinations after cross-validation. The models were evaluated on accuracy, and the best-performing combination of hyperparameters was selected.

\subsection{Training Results}

% \begin{table}[t]
% \centering
% \begin{tabular}{l|c|c|c|c|c}
% \hline \hline
% \textbf{Model}& \textbf{Accuracy} & \textbf{F1 Score} & \textbf{Recall}& \textbf{Mean Time (ms)} & \textbf{Std. Deviation (ms)} \\
% \hline
% \textbf{FNN}   & 0.9857 & 0.9764 & 0.9784& 2.70 & 0.95  \\
% \textbf{L-GBM} & 0.9875 & 0.9873 & 0.9875& 1.63 & 1.59  \\
% \textbf{RF}    & 0.9830 & 0.9826 & 0.9830& 3.90 & 0.62 \\
% \textbf{HGBT}  & 0.9792 & 0.9789 & 0.9792& 4.54 & 1.40 \\ \hline \hline
% \end{tabular}
% \caption{Evaluation metrics of the trained models.}
% \label{tab:gestures_models}
% \vspace{-6mm}
% \end{table}

\begin{table}[t]
\centering
\begin{tabular}{l|c|c|c|c|c}
\hline \hline
\multicolumn{1}{c|}{\textbf{Model}} & 
\multicolumn{1}{c|}{\textbf{Accuracy}} & 
\multicolumn{1}{c|}{\textbf{F1 Score}} & 
\multicolumn{1}{c|}{\textbf{Recall}} & 
\multicolumn{1}{c|}{\raisebox{-0.5\height}{\shortstack{\textbf{Mean}\\\textbf{(ms)}}}} & 
\multicolumn{1}{c}{\raisebox{-0.5\height}{\shortstack{\textbf{Std. Dev.}\\\textbf{(ms)}}}} \\
\hline
\textbf{FNN}   & 0.9857 & 0.9764 & 0.9784 & 2.70 & 0.95  \\
\textbf{L-GBM} & 0.9875 & 0.9873 & 0.9875 & 1.63 & 1.59  \\
\textbf{RF}    & 0.9830 & 0.9826 & 0.9830 & 3.90 & 0.62 \\
\textbf{HGBT}  & 0.9792 & 0.9789 & 0.9792 & 4.54 & 1.40 \\
\hline \hline
\end{tabular}
\caption{Evaluation metrics of the trained models.  The last two columns represent the real-time prediction time for each model (estimated over 1000 trials).}
\label{tab:gestures_models}
\vspace{-6mm}
\end{table}

Table~\ref{tab:gestures_models} shows the evaluation results on the test set. Based on the metrics, all models had analogous performance in estimating the hand gestures. Hence, we conducted an additional live test in which the models predicted the gestures in real time for 1000 iterations each. As shown in Table~\ref{tab:gestures_models}, L-GBM was the fastest among the models with an average delay of 1.63 ms. The speed is mainly due to its efficient way of growing trees and making their structure as compact as possible. The hyperparameters that were used for training the current L-GBM model are as follows: learning\_rate=0.1, max\_depth=15, min\_child\_samples=10, n\_estimators=100, and num\_leaves=15.

% The trained models have been tested on an unseen dataset, and the training of the models above (tested on a new, unseen dataset). The models were implemented in ROS and showed their performance. \lb{It is the case now!}

% for implementation due to its high prediction speed (cf. Tab \ref{tab:gestures_time}) and high accuracy, the FNN model has been rejected due to the higher number of epochs necessary to achieve similar results (1000 epochs vs 100 $n.\ estimators$ for L-GBM), this indicates that in future works other network structures need to be explored. To estimate the delay imposed by the model, we used the method described in Section~\ref{control result}, both with and without using the gesture prediction node (instead, we used the keyboard to send the inputs). The estimated delay generated by the model was 2 ms and was the smallest of all the models tested; as L-GBM, RF, and HGBT have been tested on the same grid, we did not estimate the speed with other hyperparameters that the one selected by the GridSearchCV.
% From Fig. \ref{fig:confusion}, it is noticeable that the \textbf{Pinky} and \textbf{Ring} gestures are associated with more errors. This observation can be attributed to the similarity between these two gestures.  

% \input{figures/confusion_prediction}

When using the classifiers in real-time applications, the classifier’s output is often post-processed using a majority voting technique~\cite{varol2009multiclass}. While this introduces fixed time delays depending on voting buffer size, it enhances the accuracy and reduces short-term false transitions. For our application, we set the voting buffer size to 20. Thus, the predictions of the L-GBM model over the last 20 iterations are stored in a $20\times3$ matrix, where each row contains the classifier estimations for each gesture. The gesture (column) with the highest sum is selected at the end.

% \begin{table}[t]
% \centering
% \begin{tabular}{l|c|c}
% \hline \hline
% \textbf{Model}& \textbf{Mean Time (ms)}& \textbf{Std. Deviation (ms)}  \\
% \hline
% \textbf{FNN}   & 2.70 & 0.95 \\
% \textbf{L-GBM} & 1.63 & 1.59 \\
% \textbf{RF}    & 3.90 & 0.62 \\
% \textbf{HGBT}  & 4.54 & 1.40 \\ \hline \hline
% \end{tabular}
% \caption{Real-time prediction time for each model (1000 trials).}
% \label{tab:gestures_time}
% \vspace{-5mm}
% \end{table}

\section{Experimental Results}

To evaluate our system, 5 surgeons and 2 medical doctors with experience in using the da Vinci surgical console for surgical tasks were asked to use the system to complete one of the standard da Vinci training exercises~\cite{DEROSSIS1998482} with both the console and the sensory gloves. The original exercise was modified to suit our system's single-arm setup. The modified task required participants to control the PSM with their left hand to lift a peg from one pin on the pegboard and transfer it to another pin. One trial repeats this task twice with two different pegs on the pegboard.

\subsection{Experimental Setup}

The experimental setup is shown in Fig.~\ref{fig:arm_glove_setup}. The user's hand starts in a position similar to that shown in the figure, away from the robot. The endoscope was positioned to allow a view of the pegs on the board and the tip of the instrument. The SCOPEYE smart glasses were connected to the vision tower, providing a stereo endoscopic view to the user. Participants underwent a brief training session, lasting approximately 2 to 10 minutes, primarily to familiarize themselves with the glove and the operation of the clutch mechanism. Each participant performed the exercise 5 times, both with the glove system and with the dVRK console. 

% Random motions around the workspace were executed while looking at the endoscope video from the vision tower, and clutch motions were carried out twice, during which the hand was moved at a higher speed to highlight that the robot remained stationary. Throughout the recording session, adjustments were made to the robot's jaw by varying the distances between the index and thumb tips. 

\subsection{Control Results}
\label{control result}

\begin{figure}[t]
    \centering
    \begin{subfigure}[t]{\columnwidth}
        \centering
%        \includesvg[width=\columnwidth]{glove_psm_P}
        \includegraphics[width=\columnwidth]{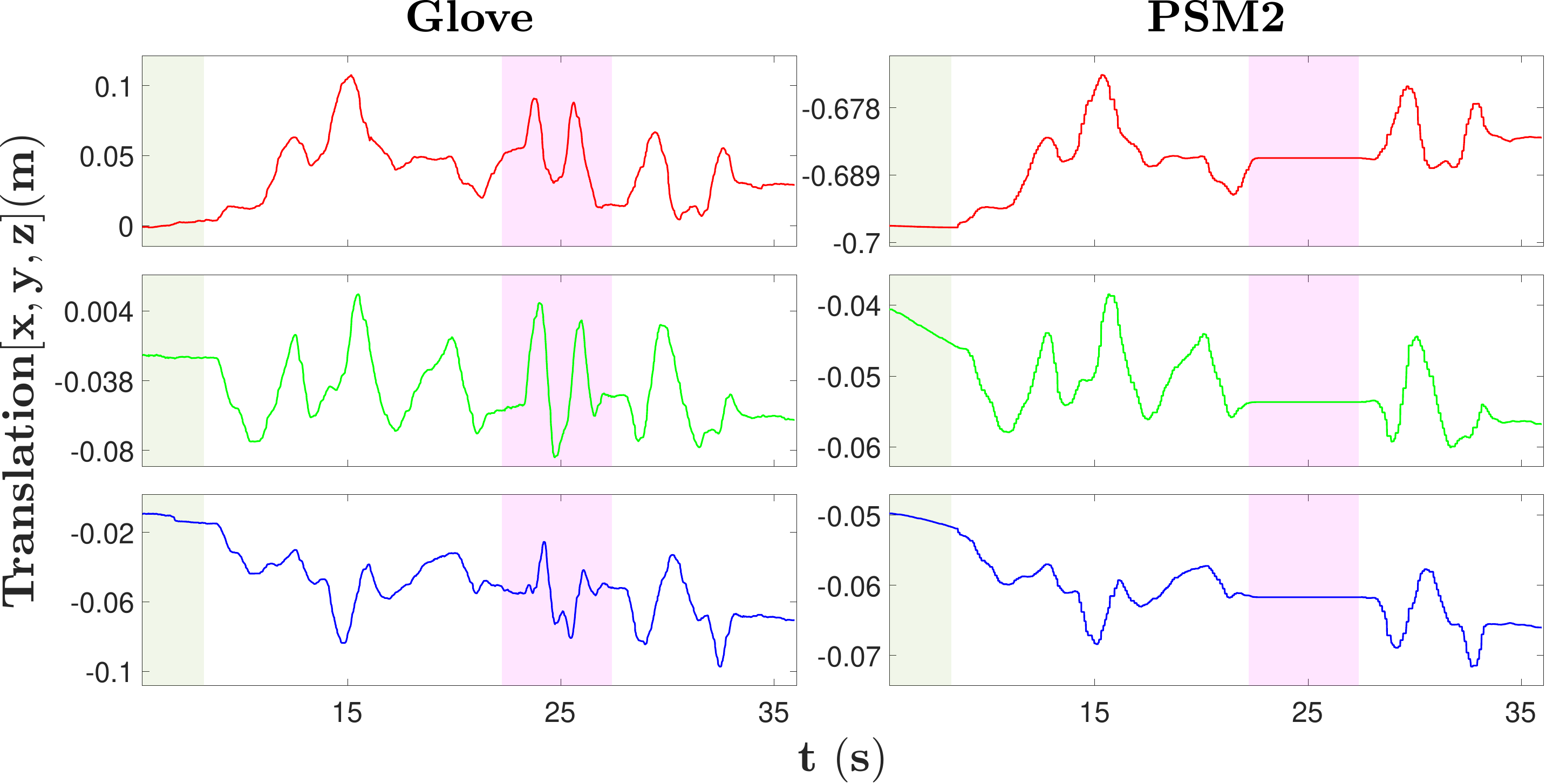}
        \caption{Translation}
        \label{fig:glove_psm_P}
    \end{subfigure}
    \hfill
    \begin{subfigure}[t]{\columnwidth}
        \centering
%        \includesvg[width=\columnwidth]{glove_psm_R}
        \includegraphics[width=\columnwidth]{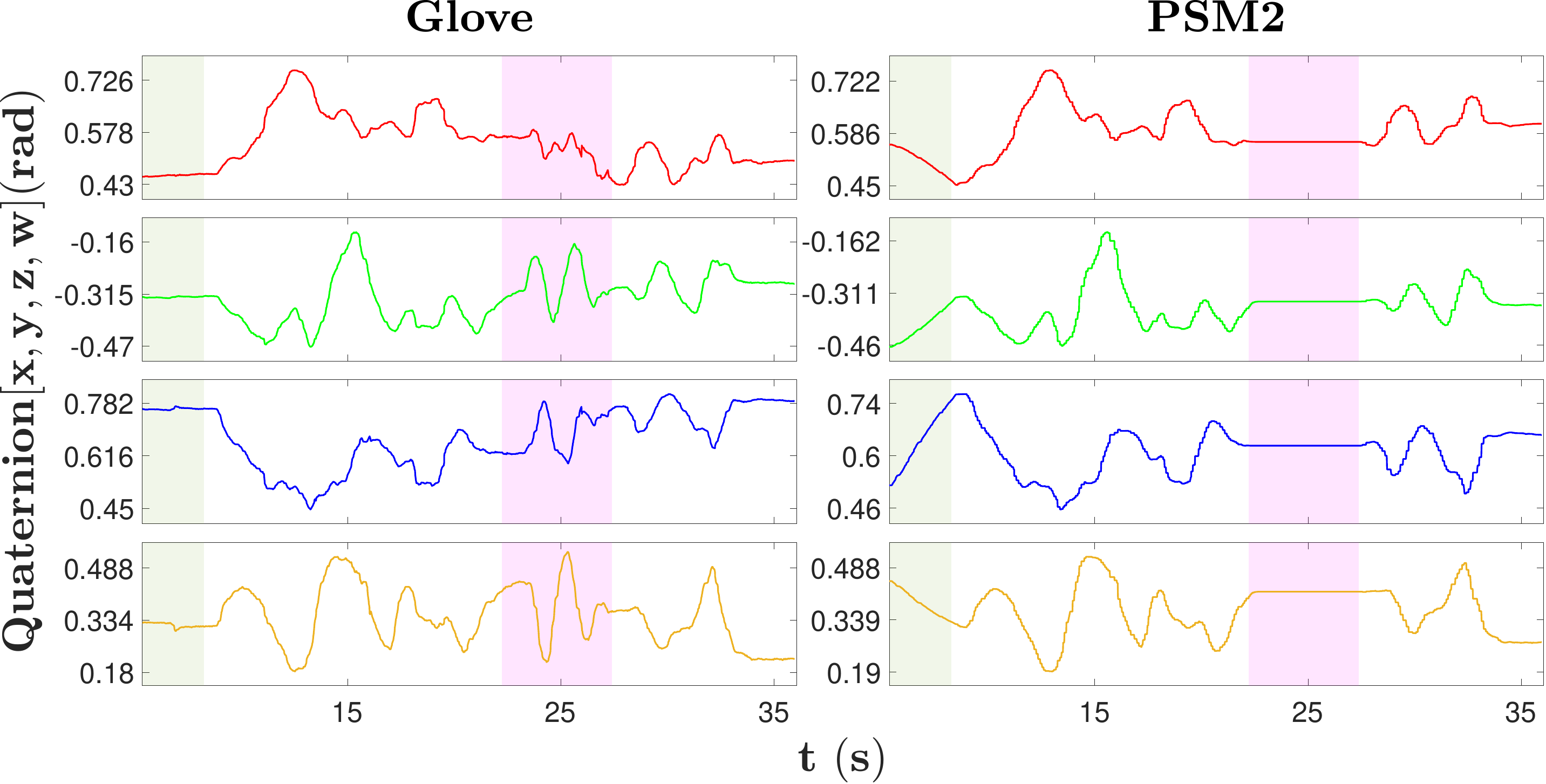}
        \caption{Orientation}
        \label{fig:glove_psm_R}
    \end{subfigure}
    \hfill
    \begin{subfigure}[t]{0.95\columnwidth}
        \centering
%        \includesvg[width=\columnwidth]{glove_psm_jaw}
        \includegraphics[width=\columnwidth]{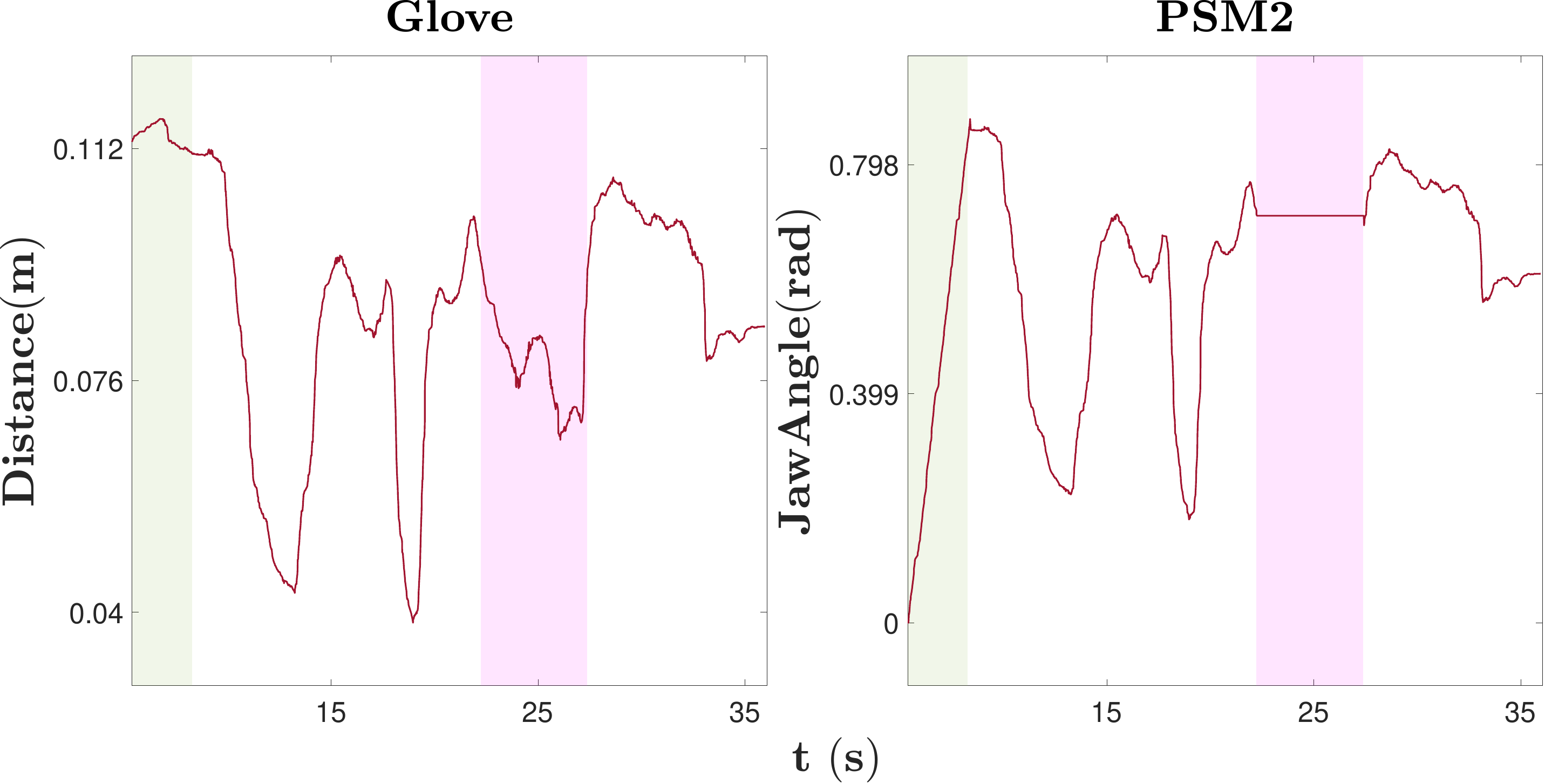}
        \caption{Jaw Movement}
        \label{fig:glove_psm_jaw}
    \end{subfigure}
    
    % \begin{subfigure}[t]{\columnwidth}
    %     \centering
    %     \includesvg[width=\columnwidth]{figures/glove_psm_tf/glove_psm_3d}
    %     \caption{3D Trajectory}
    %     \label{fig:glove_psm_3D}
    % \end{subfigure}
\caption{A sample trajectory of the glove and the PSM: (a) translation; (b) orientation; and (c) the distance of the thumb and index fingers on the glove and the jaw angle of the PSM. The gray and purple regions correspond to the periods when \textbf{Ring} and \textbf{Clutch} gestures, respectively, are active.}

%(d) focuses on a portion of the motion in translation when the clutch was disengaged, demonstrating the alignment within the 3D space.

\label{fig:glove_psm_T}
\vspace{-5mm}
\end{figure}

\begin{figure}[t!]

\centering
\includegraphics[trim={1cm 0cm 1cm 0cm}, clip, width=\columnwidth]{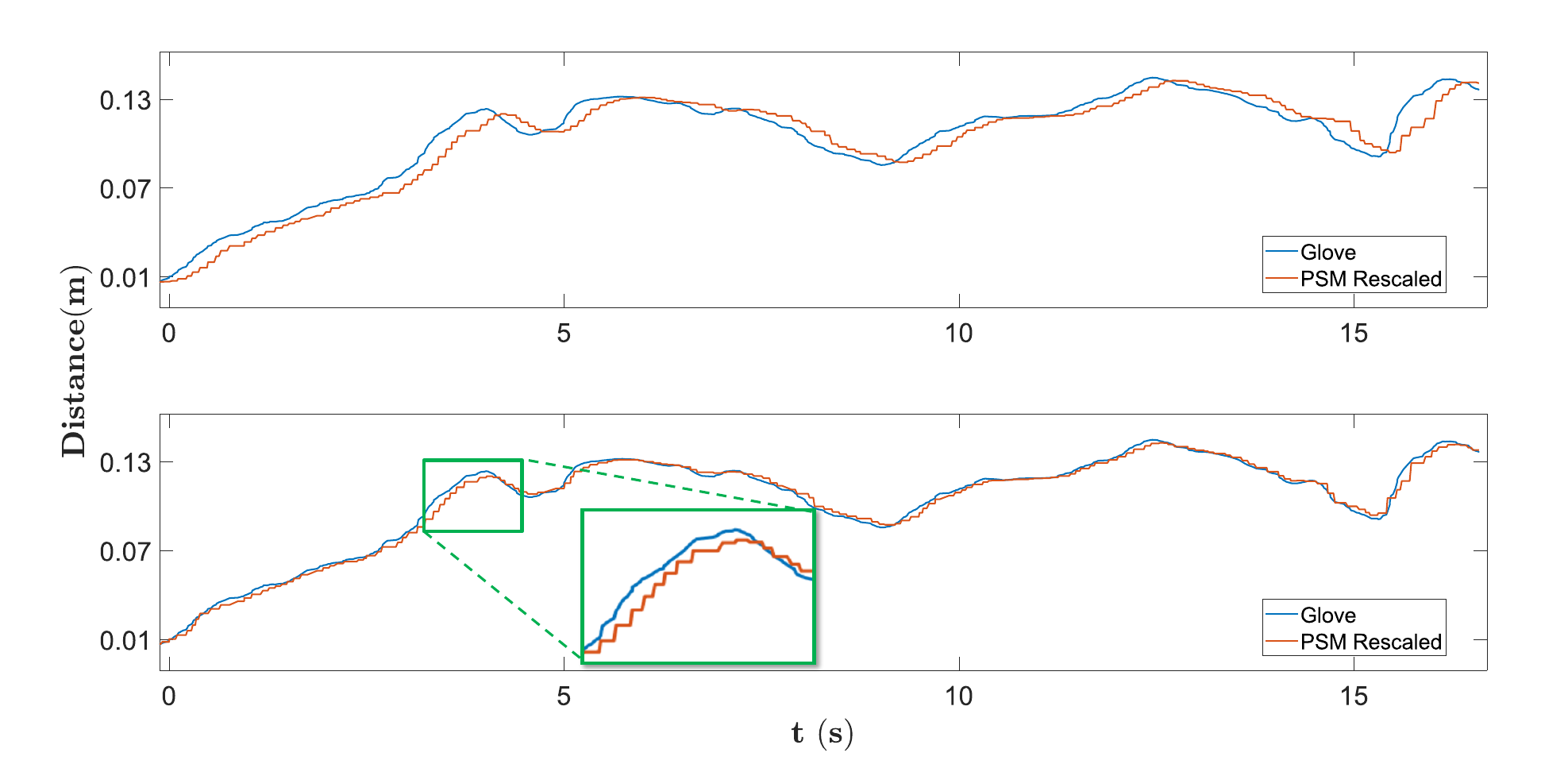}
\caption{The x-axis translation of the glove and the PSM tip: raw data (top) and data after the delay has been eliminated (bottom).}

\label{fig:glove_psm_dtw}
\vspace{-1mm}
\end{figure}

A sample trajectory of the da Vinci arm controlled by one of the surgeons with our proposed system is shown in Fig.~\ref{fig:glove_psm_T}. A comparison of the kinematics of the glove and the instrument tip is shown in Figs.~\ref{fig:glove_psm_P} and \ref{fig:glove_psm_R}, respectively. The motion of the PSM exhibits a staircase movement due to the implementation of position control of the arm in dVRK (green box in Fig.~\ref{fig:glove_psm_dtw}), where the robot stops once it reaches the reference position. 

In Fig.~\ref{fig:glove_psm_T}, the shaded region in gray (3 seconds) corresponds to the initialization of the system. This occurs after the \textbf{Ring} gesture has been activated for two seconds. During initialization, the arm tries to align with the pose of the glove.
%is the initial alignment of the PSM with the glove pose when the \textbf{Ring} gesture is held for two seconds.
The shaded region in purple corresponds to the period during which the \textbf{Clutch} was activated. The \textbf{Clutch} prevents the PSM from moving; the motion is resumed when the \textbf{Clutch} is released. The motion of the index and thumb of the glove and the instrument jaw is depicted in Fig.~\ref{fig:glove_psm_jaw}. %Currently, the fully closed configuration of the FBF jaw is represented by a negative value in the dVRK's system (as the $0^{\circ}$ angle corresponds to the jaw arms being parallel and not completely closed).
%The slight difference in extreme values between the glove and the PSM is due to the scaling between the glove and the jaw.

\begin{table}[t]
\centering
\begin{tabular}{c|cc|cc|c}
\hline
\hline
\multirow{2}{*}{\textbf{Controller}} &
  \multicolumn{2}{c|}{\textbf{\begin{tabular}[c]{@{}c@{}}Translational \\ Error (m)\end{tabular}}} &
  \multicolumn{2}{c|}{\textbf{\begin{tabular}[c]{@{}c@{}}Rotational \\ Error (rads)\end{tabular}}} &
  \multirow{2}{*}{\textbf{Delay (s)}} \\ \cline{2-5}
                 & \multicolumn{1}{c|}{\textbf{Mean}} & \textbf{Std. Dev.} & \multicolumn{1}{c|}{\textbf{Mean}} & \textbf{Std. Dev.} &       \\ \hline
\textbf{Glove}   & \multicolumn{1}{c|}{0.005}         & 0.010              & \multicolumn{1}{c|}{0.096}         & 0.146              & 0.205 \\ \hline
\textbf{Console} & \multicolumn{1}{c|}{0.002}         & 0.001              & \multicolumn{1}{c|}{0.038}         & 0.066              & 0.001 \\ \hline \hline
\end{tabular}
\caption{
The average translational and rotational errors between each controller and the PSM across all participants. The last column shows the average delay between the controller and the PSM tip signals.}
\label{tab:ctrl_precision}
\vspace{-9mm}
\end{table}
\begin{table}[ht]
\centering
\begin{tabular}{c|cc|cc}
\hline \hline
\multirow{2}{*}{\textbf{User}} & \multicolumn{2}{c|}{\textbf{Glove}} & \multicolumn{2}{c}{\textbf{Console}} \\ \cline{2-5} 
 &
  \multicolumn{1}{c|}{\textbf{\begin{tabular}[c]{@{}c@{}}Avg. Time\\ (s)\end{tabular}}} &
  \textbf{\begin{tabular}[c]{@{}c@{}}Avg. Travel\\ Distance (m)\end{tabular}} &
  \multicolumn{1}{c|}{\textbf{\begin{tabular}[c]{@{}c@{}}Avg. Time\\ (s)\end{tabular}}} &
  \textbf{\begin{tabular}[c]{@{}c@{}}Avg. Travel\\ Distance (m)\end{tabular}} \\ \hline
\color[HTML]{009901} \textbf{A}               & \multicolumn{1}{c|}{42.7} & 0.503 & \multicolumn{1}{c|}{51.9} & 0.585 \\
\color[HTML]{009901}\textbf{B}               & \multicolumn{1}{c|}{40.3} & 0.541 & \multicolumn{1}{c|}{35.8} & 0.479 \\
\color[HTML]{009901}\textbf{C}               & \multicolumn{1}{c|}{33.5} & 0.442 & \multicolumn{1}{c|}{29.2} & 0.375 \\
\color[HTML]{FE0000}\textbf{D}               & \multicolumn{1}{c|}{39.3} & 0.606 & \multicolumn{1}{c|}{18.5} & 0.298 \\
\color[HTML]{009901}\textbf{E}               & \multicolumn{1}{c|}{50.2} & 0.726 & \multicolumn{1}{c|}{30.8} & 0.365 \\
\color[HTML]{009901}\textbf{F}               & \multicolumn{1}{c|}{38.} & 0.479 & \multicolumn{1}{c|}{27.5} & 0.302 \\
\color[HTML]{FE0000}\textbf{G}               & \multicolumn{1}{c|}{34.3} & 0.503 & \multicolumn{1}{c|}{27.1} & 0.382 \\ \hline \hline
\textbf{Total}  & \multicolumn{1}{c|}{39.8} & 0.543 & \multicolumn{1}{c|}{31.6} & 0.398 \\ \hline \hline
\end{tabular}
\caption{Performance evaluation. The first column displays the average runtime, and the subsequent column shows the average traveled Euclidean distance of the instrument tip manipulated by each controller of the five trials from each participant.
}

% \lb{we have to precise the scaling factor of the console, how much it was 2.5? I forgot}}
% This is the movement of the arm, not the controllers

\label{tab:ctrl_perf}
\vspace{-5mm}
\end{table}

To compare the input motion of the glove and the output motion of the instrument tip, we first reversed the scaling (Eq.~\eqref{eq:rescale}) of the movement of the instrument tip (Fig.~\ref{fig:glove_psm_dtw}). The figure shows a delay between the two trajectories, mainly due to the computation time for the inverse kinematics. To accurately estimate the error between the input and output signals, we identified the delay by the peak positions within each signal and eliminated the effect of the delay by shifting the PSM signal to align with the glove's signal (Fig.~\ref{fig:glove_psm_dtw}). The difference in translation was quantified using the Euclidean distance, while the discrepancy in rotation was determined using the standard bi-invariant metric on the group of rotations~\cite{zefran_metrics_1999}. The errors for each controller in all the trials across the participants are listed in Table~\ref{tab:ctrl_precision}. Although the magnitude of the errors is larger for our system, the task completion times in the two setups were comparable (Table~\ref{tab:ctrl_perf}), suggesting that the errors were acceptable. These larger errors might also explain the slightly higher average travel distances for our system.

The main reason for significantly higher std. deviation for our system is mainly the consequence of safety mechanisms in the dVRK console: when the difference between the PSM position and the console position is larger than a threshold (as when the PSM is held by a peg, for example), the system shuts down. No such mechanism is employed in our system, leading to the presence of large errors.

The average delay of our system is $0.2$s, significantly larger than for the dVRK console. However, \cite{woodsFactorsInfluencingLatency2015} shows that such delays are routinely tolerated by humans.

We evaluated the average performance of the participants with each controller (Table~\ref{tab:ctrl_perf}). Performance is evaluated by the average duration and travel distance of the instrument tip, indicating how efficiently they managed to perform the task. Considering that this is a prototype system, the performance of controlling the robot with the glove compared favorably with that of the console.

% The errors derived from Figs.~\ref{fig:glove_psm_P} and \ref{fig:glove_psm_R} are shown in . The duration and error metrics for each participant's performance are depicted in Table~\ref{tab:ctrl_precision}. The average delay in all trials and all participants was 0.223s. 

\subsection{Survey Results}

Participants were asked to compare the usability and task load of our proposed system with the conventional console. Usability was assessed with the System Usability Scale (SUS)~\cite{SUS}, and task load was measured using NASA Task Load Index (TLX)~\cite{nasa_questionnaire} questionnaires. The overall results are shown in Table~\ref{tab:SUS_NTX}.

The average SUS score for the glove was 67.5 compared to the console's 81.4 (13.9 points of difference). These results align well with the quantitative assessment of the performance in Table~\ref{tab:ctrl_perf}: users who achieved similar performance between the glove and the console responded with a similar SUS assessment for the two. Considering that users have significantly more training on the dVRK console and that our system is just a prototype, we judge the reported SUS score as overwhelmingly positive.  

% For instance, User G rated the glove controller's SUS score at 3.3, almost matching their 3.7 score for the console. This demonstrates that, even without advanced hardware, the glove offers a user experience that is well-received by professionals. 
When considering the workload, the average TLX score of the glove was higher than that of the console (40.6 vs. 26.1). We hypothesize that this is mainly due to the lack of ergonomic support in our system, while the dVRK console is equipped with a proper armrest and headrest. In addition, as in the case of the SUS scores, users are more familiar with the da Vinci console setup, significantly reducing the effort needed to adapt to the system. Thus, we anticipate that these scores would improve if users spent more time with our system.

\section{Conclusion}

This paper presents a novel surgeon interface for the da Vinci surgical robot. The system integrates the Manus Meta Prime 3 Haptic XR glove, the HTC Vive Tracker, and SCOPEYE wireless smart glasses and controls one arm of the dVRK surgical robot platform. In addition to moving the arm, the surgeon can use fingers to control the end-effector of the surgical instrument. A combination of scaling and gesture recognition was used to achieve precise and intuitive control of the robot's movements while implementing an enhanced version of the clutch mechanism. In particular, we introduce clutching of the instrument orientation, a functionality unavailable in the da Vinci system. The vibrotactile feedback of the glove is used to alert the user when gesture commands are invoked. 

The proposed interface offers several potential advantages. Surgeons can directly control robotic arms by moving their hands without the need for additional physical devices. This facilitates a more immersive surgeon-robot interaction. However, it should be noted that the hardware used in our setup is not the latest available on the market; we used the first version of HTC Vive Trackers and Manus Meta Prime 3 XR, both of which have lower precision compared to the next generation devices (HTC Vive Trackers v2 and Quantum XR Metaglove).

\begin{table}[t]
\centering
\begin{tabular}{c|cccc}
\hline \hline
                                  & \multicolumn{4}{c}{\textbf{Controller}}                                                                                  \\ \cline{2-5} 
                                  & \multicolumn{2}{c|}{\textbf{Glove}}                                   & \multicolumn{2}{c}{\textbf{Console}}             \\ \cline{2-5} 
\multirow{-3}{*}{\textbf{User}}   & \multicolumn{1}{c|}{\textbf{SUS}} & \multicolumn{1}{c|}{\textbf{TLX}} & \multicolumn{1}{c|}{\textbf{SUS}} & \textbf{TLX} \\ \hline
{\color[HTML]{009901} \textbf{A}} & \multicolumn{1}{c|}{92.5}  &  \multicolumn{1}{c|}{27  }                & \multicolumn{1}{c|}{92.5} & 35                    \\
{\color[HTML]{009901} \textbf{B}} & \multicolumn{1}{c|}{70}  &  \multicolumn{1}{c|}{45  }                & \multicolumn{1}{c|}{85} & 27                    \\
{\color[HTML]{009901} \textbf{C}} & \multicolumn{1}{c|}{67.5}  &  \multicolumn{1}{c|}{45  }                & \multicolumn{1}{c|}{60} & 65.8                  \\
{\color[HTML]{FE0000} \textbf{D}} & \multicolumn{1}{c|}{80}  &  \multicolumn{1}{c|}{10.8}                & \multicolumn{1}{c|}{80} & 5                     \\
{\color[HTML]{009901} \textbf{E}} & \multicolumn{1}{c|}{22.5}  &  \multicolumn{1}{c|}{60.8}                & \multicolumn{1}{c|}{92.5} & 11.6                  \\
{\color[HTML]{009901} \textbf{F}} & \multicolumn{1}{c|}{82.5}  &  \multicolumn{1}{c|}{48  }                & \multicolumn{1}{c|}{92.5} & 12.5                  \\
{\color[HTML]{FE0000} \textbf{G}} & \multicolumn{1}{c|}{57.5}  &  \multicolumn{1}{c|}{47  }                & \multicolumn{1}{c|}{67.5} & 25                    \\ \hline \hline
\textbf{Total Average}            & \multicolumn{1}{c|}{67.5}  &  \multicolumn{1}{c|}{40.6}                & \multicolumn{1}{c|}{81.4} & 26.1                  \\ \hline \hline
\end{tabular}
\caption{System Usability Scale (SUS) and NASA Task Load Index (TLX) questionnaire results, answered by surgeons (green letters) and medical doctors (red letters). In the SUS, a score closer to 100 indicates better usability, and in the TLX, a score closer to 100 indicates a higher workload.}
\label{tab:SUS_NTX}

\vspace{-5mm}
\end{table}

The system was tested by users who were experienced with the da Vinci system, including five surgeons and two medical doctors. They were asked to perform a surgical training task similar to those used in training with the da Vinci devices. The evaluation shows that the system is responsive and precise. Surgeons could effectively complete training exercises with minimal practice with the new interface, suggesting that the interface is highly intuitive. After the exercise, each participant completed two questionnaires to assess the usability and workload of the proposed system. Users rated their experience favorably; however, a higher workload suggests the need for a more ergonomic setup.

The existing surgical consoles are bulky and take up valuable space in the operating room. They limit the ability to optimize the operating room setup for throughput. They also pose challenges to the communication of the surgical team, leading to potential workflow disruptions. In contrast, our proposed system is inexpensive (estimated total cost of ~\$5000), as it uses off-the-shelf components. Finally it  dramatically reduces space requirements and is easily portable. The dimensions of the da Vinci S/Si console are  166 cm x 163 cm x 163 cm and it weighs 273 kg, compared to our setup including 2 HTCVive tracker (7cm × 8cm × 5cm, 100g), 2 basestations (9cm x 6cm x 8cm 0.45 kg ), 2 SCOPEYE wireless input transmitter (in total 39cm x 30cm x 16cm) two sensory gloves (worn by the surgeon, 138g) and a computer. It demonstrates how modalities such as vibrotactile feedback can efficiently provide additional information to the surgeon without cluttering the visual field. Finally, the proposed architecture allows for rapid prototyping and opens opportunities for further innovations in the design of surgical robot interfaces.

Future work includes implementing velocity control to enhance the user experience, enabling full control of both surgical arms, upgrading hardware to improve tracking precision, and expanding the repertoire of user gestures to increase the range of available actions. We will also implement enhanced ergonomic features, such as adjustable armrests and customizable control environments.

\clearpage
\nocite{*}
\bibliographystyle{IEEEtran}
\bibliography{iros_2024}

\begin{thebibliography}{10}
\providecommand{\url}[1]{#1}
\csname url@samestyle\endcsname
\providecommand{\newblock}{\relax}
\providecommand{\bibinfo}[2]{#2}
\providecommand{\BIBentrySTDinterwordspacing}{\spaceskip=0pt\relax}
\providecommand{\BIBentryALTinterwordstretchfactor}{4}
\providecommand{\BIBentryALTinterwordspacing}{\spaceskip=\fontdimen2\font plus
\BIBentryALTinterwordstretchfactor\fontdimen3\font minus \fontdimen4\font\relax}
\providecommand{\BIBforeignlanguage}[2]{{%
\expandafter\ifx\csname l@#1\endcsname\relax
\typeout{** WARNING: IEEEtran.bst: No hyphenation pattern has been}%
\typeout{** loaded for the language `#1'. Using the pattern for}%
\typeout{** the default language instead.}%
\else
\language=\csname l@#1\endcsname
\fi
#2}}
\providecommand{\BIBdecl}{\relax}
\BIBdecl

\bibitem{kanji2021room}
F.~Kanji, T.~Cohen, M.~Alfred, A.~Caron, S.~Lawton, S.~Savage, D.~Shouhed, J.~T. Anger, and K.~Catchpole, ``Room size influences flow in robotic-assisted surgery,'' \emph{International Journal of Environmental Research and Public Health}, vol.~18, no.~15, p. 7984, 2021.

\bibitem{bharathan2013operating}
R.~Bharathan, R.~Aggarwal, and A.~Darzi, ``Operating room of the future,'' \emph{Best Practice \& Research Clinical Obstetrics \& Gynaecology}, vol.~27, no.~3, pp. 311--322, 2013.

\bibitem{sandberg2005deliberate}
W.~S. Sandberg, B.~Daily, M.~Egan, J.~E. Stahl, J.~M. Goldman, R.~A. Wiklund, and D.~Rattner, ``Deliberate perioperative systems design improves operating room throughput,'' \emph{The Journal of the American Society of Anesthesiologists}, vol. 103, no.~2, pp. 406--418, 2005.

\bibitem{stahl2006reorganizing}
J.~E. Stahl, W.~S. Sandberg, B.~Daily, R.~Wiklund, M.~T. Egan, J.~M. Goldman, K.~B. Isaacson, S.~Gazelle, and D.~W. Rattner, ``Reorganizing patient care and workflow in the operating room: a cost-effectiveness study,'' \emph{Surgery}, vol. 139, no.~6, pp. 717--728, 2006.

\bibitem{simorov_review_2012}
A.~Simorov, R.~S. Otte, C.~M. Kopietz, and D.~Oleynikov, ``Review of surgical robotics user interface: What is the best way to control robotic surgery?'' \emph{Surgical Endoscopy}, vol.~26, no.~8, pp. 2117--2125, Aug. 2012.

\bibitem{cofran2021barriers}
L.~Cofran, T.~Cohen, M.~Alfred, F.~Kanji, E.~Choi, S.~Savage, J.~Anger, and K.~Catchpole, ``Barriers to safety and efficiency in robotic surgery docking,'' \emph{Surgical endoscopy}, pp. 1--10, 2021.

\bibitem{hong_head-mounted_2019}
N.~Hong, M.~Kim, C.~Lee, and S.~Kim, ``Head-mounted interface for intuitive vision control and continuous surgical operation in a surgical robot system,'' \emph{Medical \& Biological Engineering \& Computing}, vol.~57, no.~3, pp. 601--614, Mar. 2019.

\bibitem{wen_hand_2014}
R.~Wen, W.-L. Tay, B.~P. Nguyen, C.-B. Chng, and C.-K. Chui, ``Hand gesture guided robot-assisted surgery based on a direct augmented reality interface,'' \emph{Computer Methods and Programs in Biomedicine}, vol. 116, no.~2, pp. 68--80, Sep. 2014.

\bibitem{Ai2024}
L.~Ai, P.~Kazanzides, and E.~Azimi, ``Mixed reality based teleoperation and visualization of surgical robotics,'' \emph{Healthcare Technology Letters}, vol.~11, no. 2-3, pp. 179--188, Mar 2024.

\bibitem{intuitive2023catalog}
{Intuitive}, ``Intuitive da vinci x/xi system instrument and accessory catalog,'' \url{https://www.intuitive.com/en-us/-/media/ISI/Intuitive/Pdf/da-vinci-x-xi-instruments-accessories-catalog.pdf}, Dec. 2023, accessed: 2025-03-03.

\bibitem{mercilinraajini2023mem}
X.~MercilinRaajini, S.~Abirami, S.~Keerthana, N.~Shrivastava, N.~Malokhat, and R.~Yokeshwaran, ``Mem based hand gesture controlled wireless robot,'' in \emph{E3S Web of Conferences}, vol. 399.\hskip 1em plus 0.5em minus 0.4em\relax EDP Sciences, 2023, p. 01013.

\bibitem{burns2020design}
M.~K. Burns and R.~Vinjamuri, ``Design of a soft glove-based robotic hand exoskeleton with embedded synergies,'' \emph{Advances in Motor Neuroprostheses}, pp. 71--87, 2020.

\bibitem{brygo2017synergy}
A.~Brygo, I.~Sarakoglou, G.~Grioli, and N.~Tsagarakis, ``Synergy-based bilateral port: A universal control module for tele-manipulation frameworks using asymmetric master--slave systems,'' \emph{Frontiers in bioengineering and biotechnology}, vol.~5, p.~19, 2017.

\bibitem{htcvive}
\BIBentryALTinterwordspacing
{HTC Vive}, \emph{HTC Vive Tracker Developer Guidelines}.\hskip 1em plus 0.5em minus 0.4em\relax HTC Vive, 2020. [Online]. Available: \url{https://developer.vive.com/documents/721/HTC-Vive-Tracker-2018-Developer-Guidelines_v1.5.pdf}
\BIBentrySTDinterwordspacing

\bibitem{manus}
\BIBentryALTinterwordspacing
{MANUS}, \emph{MANUS Prime 3 Haptic XR full spec sheet}.\hskip 1em plus 0.5em minus 0.4em\relax MANUS, 2021. [Online]. Available: \url{https://assets.website-files.com/61de97d15a7bb6441d9565c0/65127f073666b95695852c32_Datasheet%20-%20Prime%203%20Haptic%20XR%20(small).pdf}
\BIBentrySTDinterwordspacing

\bibitem{dvrk}
P.~Kazanzides, Z.~Chen, A.~Deguet, G.~S. Fischer, R.~H. Taylor, and S.~P. DiMaio, ``An open-source research kit for the da vinci surgical system,'' in \emph{IEEE Intl. Conf. on Robotics and Auto. (ICRA)}, Hong Kong, China, 2014, pp. 6434--6439.

\bibitem{scopeye}
\BIBentryALTinterwordspacing
{MediThinkQ}, \emph{SCOPEYE Catalog}.\hskip 1em plus 0.5em minus 0.4em\relax MediThinkQ, 2023. [Online]. Available: \url{https://s3.ap-northeast-2.amazonaws.com/logicsquare-seoul/e3a824df-6e8a-448c-b6f6-ecfe090a666f/SCOPEYE%20NEW%20BAND_Catalog_ENG_v1.0.pdf}
\BIBentrySTDinterwordspacing

\bibitem{SUS}
J.~R. Lewis, ``The system usability scale: past, present, and future,'' \emph{International Journal of Human--Computer Interaction}, vol.~34, no.~7, pp. 577--590, 2018.

\bibitem{nasa_questionnaire}
S.~G. Hart, ``Nasa-task load index (nasa-tlx); 20 years later,'' in \emph{Proceedings of the human factors and ergonomics society annual meeting}, vol.~50, no.~9.\hskip 1em plus 0.5em minus 0.4em\relax Sage publications Sage CA: Los Angeles, CA, 2006, pp. 904--908.

\bibitem{korayem2021controlling}
M.~Korayem, M.~Madihi, and V.~Vahidifar, ``Controlling surgical robot arm using leap motion controller with kalman filter,'' \emph{Measurement}, vol. 178, p. 109372, 2021.

\bibitem{ros}
\BIBentryALTinterwordspacing
{Open Source Robotics Foundation}, ``Robot operating system.'' [Online]. Available: \url{https://www.ros.org}
\BIBentrySTDinterwordspacing

\bibitem{oh2023framework}
\BIBentryALTinterwordspacing
K.-H. Oh, L.~Borgioli, M.~Zefran, L.~Chen, and P.~C. Giulianotti, ``A framework for automated dissection along tissue boundary,'' 2023. [Online]. Available: \url{https://arxiv.org/abs/2310.09669}
\BIBentrySTDinterwordspacing

\bibitem{basestation}
O.~Kreylos, ``Lighthouse tracking examined,'' \emph{Doc-ok. org}, vol.~25, p.~12, 2016.

\bibitem{Htc}
D.~C. Niehorster, L.~Li, and M.~Lappe, ``The accuracy and precision of position and orientation tracking in the htc vive virtual reality system for scientific research,'' \emph{i-Perception}, vol.~8, no.~3, p. 2041669517708205, 2017.

\bibitem{bazarevsky2019mediapipe}
V.~Bazarevsky, Y.~Golubev, Y.~Kartynnik \emph{et~al.}, ``Mediapipe: A framework for building perception pipes,'' \emph{arXiv preprint arXiv:1906.08172}, 2019.

\bibitem{hand_percentile}
D.~Daruis, N.~Khamis, and B.~Deros, ``The hand--the basic anthropometry,'' \emph{Hum. Factors Ergo J}, vol.~6, pp. 49--55, 2021.

\bibitem{breiman2001random}
L.~Breiman, ``Random forests,'' \emph{Machine learning}, vol.~45, pp. 5--32, 2001.

\bibitem{Goodfellow-et-al-2016}
I.~Goodfellow, Y.~Bengio, and A.~Courville, \emph{Deep Learning}.\hskip 1em plus 0.5em minus 0.4em\relax MIT Press, 2016, \url{http://www.deeplearningbook.org}.

\bibitem{rysbek2023recognizing}
Z.~Rysbek, K.-H. Oh, and M.~Zefran, ``Recognizing intent in collaborative manipulation,'' in \emph{Proceedings of the 25th International Conference on Multimodal Interaction}, 2023, pp. 498--506.

\bibitem{NIPS2017_6449f44a}
\BIBentryALTinterwordspacing
G.~Ke, Q.~Meng, T.~Finley, T.~Wang, W.~Chen, W.~Ma, Q.~Ye, and T.-Y. Liu, ``Lightgbm: A highly efficient gradient boosting decision tree,'' in \emph{Advances in Neural Information Processing Systems}, I.~Guyon, U.~V. Luxburg, S.~Bengio, H.~Wallach, R.~Fergus, S.~Vishwanathan, and R.~Garnett, Eds., vol.~30.\hskip 1em plus 0.5em minus 0.4em\relax Curran Associates, Inc., 2017. [Online]. Available: \url{https://proceedings.neurips.cc/paper_files/paper/2017/file/6449f44a102fde848669bdd9eb6b76fa-Paper.pdf}
\BIBentrySTDinterwordspacing

\bibitem{XGBoost}
\BIBentryALTinterwordspacing
T.~Chen and C.~Guestrin, ``{XGBoost}: A scalable tree boosting system,'' in \emph{Proceedings of the 22nd ACM SIGKDD International Conference on Knowledge Discovery and Data Mining}, ser. KDD '16.\hskip 1em plus 0.5em minus 0.4em\relax New York, NY, USA: ACM, 2016, pp. 785--794. [Online]. Available: \url{http://doi.acm.org/10.1145/2939672.2939785}
\BIBentrySTDinterwordspacing

\bibitem{friedman2002stochastic}
J.~H. Friedman, ``Stochastic gradient boosting,'' \emph{Computational statistics \& data analysis}, vol.~38, no.~4, pp. 367--378, 2002.

\bibitem{GridCV}
S.~M. LaValle, M.~S. Branicky, and S.~R. Lindemann, ``On the relationship between classical grid search and probabilistic roadmaps,'' \emph{The International Journal of Robotics Research}, vol.~23, no. 7-8, pp. 673--692, 2004.

\bibitem{varol2009multiclass}
H.~A. Varol, F.~Sup, and M.~Goldfarb, ``Multiclass real-time intent recognition of a powered lower limb prosthesis,'' \emph{IEEE Transactions on Biomedical Engineering}, vol.~57, no.~3, pp. 542--551, 2009.

\bibitem{DEROSSIS1998482}
\BIBentryALTinterwordspacing
A.~M. Derossis, G.~M. Fried, M.~Abrahamowicz, H.~H. Sigman, J.~S. Barkun, and J.~L. Meakins, ``Development of a model for training and evaluation of laparoscopic skills 11this work was supported by an educational grant from united states surgical corporation (auto suture canada).'' \emph{The American Journal of Surgery}, vol. 175, no.~6, pp. 482--487, 1998. [Online]. Available: \url{https://www.sciencedirect.com/science/article/pii/S0002961098000804}
\BIBentrySTDinterwordspacing

\bibitem{zefran_metrics_1999}
M.~{\v Z}efran, V.~Kumar, and C.~Croke, ``Metrics and {{Connections}} for {{Rigid-Body Kinematics}},'' \emph{The International Journal of Robotics Research}, vol.~18, no.~2, pp. 242--1, Feb. 1999.

\bibitem{woodsFactorsInfluencingLatency2015}
D.~L. Woods, J.~M. Wyma, E.~W. Yund, T.~J. Herron, and B.~Reed, ``\BIBforeignlanguage{en}{Factors influencing the latency of simple reaction time},'' \emph{\BIBforeignlanguage{en}{Frontiers in Human Neuroscience}}, vol.~9, Mar. 2015.

\end{thebibliography}

%%%%%%%%%%%%%%%%%%%%%%%%%%%%%%%%%%%%%%%%%%%%%%%%%%%%%%%%%%%%%%%%%%%%%%%%%%%%%%%%

\end{document}